\documentclass[10pt,leqno]{amsart}
\usepackage{graphicx}
\usepackage{indentfirst,csquotes}

\usepackage{amssymb,amsthm,amsmath}
\usepackage{xcolor,paralist,hyperref,fancyhdr,etoolbox}

\usepackage{multirow}
\usepackage{multicol}
\usepackage{caption}
\usepackage{threeparttable}
\usepackage{algorithm}
\usepackage{algpseudocode}
\usepackage[T1]{fontenc}
\usepackage{lmodern}

\usepackage{amsmath, amssymb, algorithm, algorithmicx, algpseudocode}

\newcommand{\beqs}{\vspace{0mm}\begin{eqnarray}}
\newcommand{\eeqs}{\vspace{0mm}\end{eqnarray}}
\newcommand{\barr}{\begin{array}}
\newcommand{\earr}{\end{array}}

\newcommand{\swap}

\usepackage[numbers]{natbib}
\usepackage{booktabs}

\hypersetup{ colorlinks=true, linkcolor=black, filecolor=black, urlcolor=black }

\usepackage{lipsum}

\title{Meta-SurDiff: Classification Diffusion Model Optimized by Meta Learning is Reliable for Online Surgical Phase Recognition}

\author{
Yufei Li\textsuperscript{1},
Jirui Wu\textsuperscript{1},
Long Tian\textsuperscript{1,*},
LiMing Wang\textsuperscript{1},
Xiaonan Liu\textsuperscript{2},
Zijun Liu\textsuperscript{2},
Xiyang Liu\textsuperscript{1}
}

\thanks{\textsuperscript{1} School of Computer Science and Technology, Xidian University, Xi'an, China}
\thanks{\textsuperscript{2} The First Affiliated Hospital of Air Force Military Medical University, Xi'an, China}
\thanks{*Corresponding author: Long Tian (email@example.com)}

\date{\today}
\begin{document}


\begin{abstract}
Online surgical phase recognition has drawn great attention most recently due to its potential downstream applications closely related to human life and health. 
Despite deep models have made significant advances in capturing the discriminative long-term dependency of surgical videos to achieve improved recognition, they rarely account for exploring and modeling the uncertainty in surgical videos, which should be crucial for reliable online surgical phase recognition. 
We categorize the sources of uncertainty into two types, frame ambiguity in videos and unbalanced distribution among surgical phases, which are inevitable in surgical videos. To address this pivot issue, we introduce a meta-learning-optimized classification diffusion model (Meta-SurDiff), to take full advantage of the deep generative model and meta-learning in achieving precise frame-level distribution estimation for reliable online surgical phase recognition. For coarse recognition caused by ambiguous video frames, we employ a classification diffusion model to assess the confidence of recognition results at a finer-grained frame-level instance. For coarse recognition caused by unbalanced phase distribution, we use a meta-learning based objective to learn the diffusion model, thus enhancing the robustness of classification boundaries for different surgical phases.
We establish effectiveness of Meta-SurDiff in online surgical phase recognition through extensive experiments on five widely used datasets using more than four practical metrics. The datasets include Cholec80, AutoLaparo, M2Cai16, OphNet, and NurViD, where OphNet comes from ophthalmic surgeries, NurViD is the daily care dataset, while the others come from laparoscopic surgeries. We will release the code upon acceptance.
\end{abstract} 
\maketitle

\section{Introduction}
\label{sec:intro}

Surgical phases recognition aims to identify the representation of high-level surgical stages depicted in surgical videos \citep{jin2017sv}. This capability holds potential applications for fruitful downstream tasks, such as 
automatic indexing of surgical video databases \citep{twinanda2016endonet},
real-time monitoring of surgical procedures \citep{bricon2007context}, optimizing surgeons schedules \citep{neumuth2017surgical}, evaluating surgeons' proficiency \citep{liu2021towards}, etc. 
The primary objective of surgical phase recognition is to predict the category variable $ y \in \mathbb{R}^{L \times C}$ given a video frame $ x \in \mathbb{R}^{L \times I}$, where $L$ and $C$ denote the video frame length and category of phase number, $I$ is the channel number per frame.
The process is characterized by the deterministic function $f(x) \in \mathbb{R}^{L \times C}$ that transforms the video frame $x$ into the category variable $y$.To help alert surgeons and support decision-making in real-time during surgery, we do not use the future information within the video frame of $x$, which is also known as online phase recognition \citep{quellec2014real,dergachyova2016automatic}. It requires us to design the mapping function $f(\cdot)$ carefully without the information leakage. 

\begin{figure}[!t]
\begin{center}
\centerline{\includegraphics[width=0.6\textwidth]{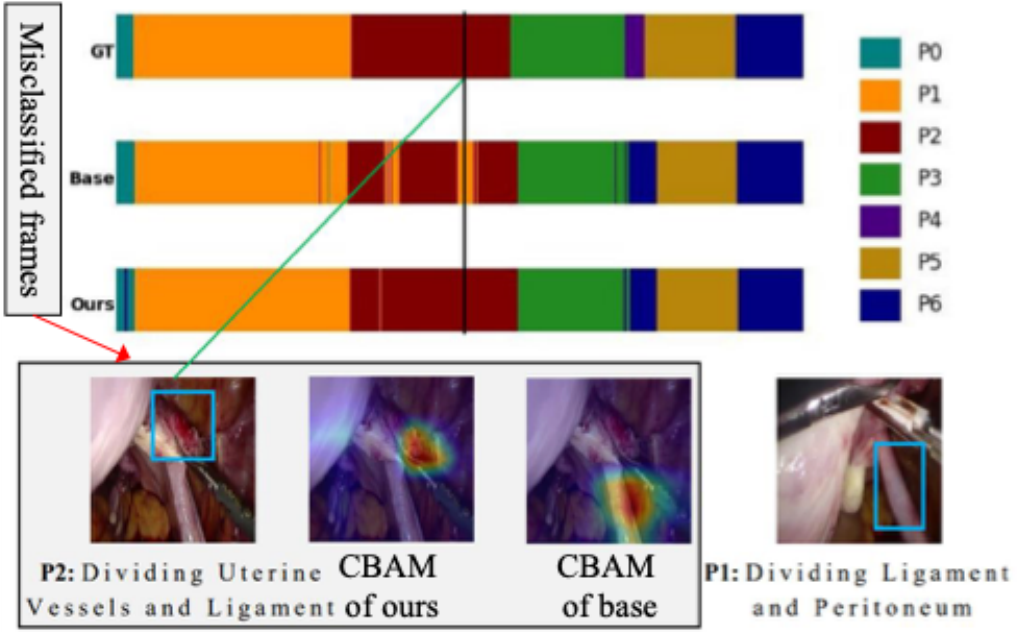}}
\caption{The illustration of unbalanced phase distribution and frame ambiguity on AutoLaparo dataset. i) Unbalanced phase distribution: The ribbon charts tell us that the frame distribution across different phases (best viewed in colors) is highly unbalanced. ii) Frame ambiguity: The blue box inside the black box indicates the target organ and tool that should be focused on while the blue box outside the black box represents incorrectly focused area. The correct and incorrect areas are visual similar which might confuse the model during recognition. Along with the CBAM results, we believe that a better approach to address the aforementioned issues would be beneficial to reliable OSP recognition.
}
\label{fig:motivation}
\end{center}
\vspace{-8mm}
\end{figure}

Deep neural network-based models \citep{he2016deep,vaswani2017attention} have shown promising performance in surgical phase recognition by designing complex deterministic functions $f(\cdot)$\citep{jin2021temporal,liu2023skit,ding2022exploring,zhao2022real,rivoir2024pitfalls,FENG2024103026, ZHANG2021102224,JIN2020101572}. To capture long-term spatial information, Transformer-based methods \citep{tao2023last,yue2023cascade} have been introduced, where SKiT \citep{liu2023skit} improving efficiency via key pooling ($\mathcal{O}(1)$ complexity) while preventing future information leakage. Recently, CNN-based models have regained attention due to batch normalization pitfalls \citep{rivoir2024pitfalls}. To reduce the labor of video annotation, UATD \citep{ding2022exploring} and VTDC \citep{shi2021semi} utilize timestamp annotation and semi-supervised learning.
\textcolor{black}{However, previous methods neglect the impact of inherent frame ambiguity and unbalanced phase distribution in surgical videos on the robustness of Online Surgical Phase (OSP) recognition, as shown in Fig. \ref{fig:motivation}. For example, in laparoscopic rectal cancer surgery, the high similarity between the sigmoid colon mobilization and microvascular mobilization, along with irregular camera angle changes caused by emergencies during the surgery, contributes to ambiguity in the surgical video frames. Additionally, free rectal movements are much more frequent than other movements because they are central to the procedure. Conversely, digestive tract reconstruction action occurs infrequently due to their fixed process, resulting in unbalanced phase distribution. Therefore, overlooking these inherent factors in surgical videos during phase recognition may lead to unforeseen suboptimal outcomes and significantly misguide downstream surgical tasks. 
}

In this paper, we propose a meta-learning optimized classification diffusion model for reliable OSP recognition (Meta-SurDiff) to address the aforementioned issues in a unified end-to-end framework. 
Our primary objective is to mitigate the negative impact of frame ambiguity on OSP recognition, conditioned on some coarse phase representations derived from any available deep models mentioned above. To achieve this goal, we introduce a novel classification diffusion model, which unifies the conditional diffusion generative process \citep{song2020denoising,ho2020denoising} with coarse phase representations. Notably, to demonstrate the superiority of Meta-SurDiff, we use ConvNext+LSTM, a simple yet effective backbone \citep{rivoir2024pitfalls}, to extract the phase representations.
Although the classification diffusion model should be robust for frame ambiguity, the model still faces the risk of being biased towards the majority phases due to the unbalanced distribution of frames. To this end, we introduce a re-weighting based meta-learning objective to balance the negative impact of unbalanced phases on optimizing the model.
We summarize the main contributions as follows: 
\begin{itemize}
    \item In the realm of OSP recognition, we introduce Meta-SurDiff, a meta-learning optimized classification diffusion model. It considers the covariate-dependence across both the forward and reverse processes within the diffusion model based on coarse phase representations, resulting in a highly accurate phase distribution estimation.
    \item Meta-SurDiff serves as a flexible plugin framework, seamlessly campatible with existing well-designed models for OSP recognition, using their strong capability to estimate the coarse phase representations, facilitating the estimation of complete phase distribution for reliable recognition.
    \item Experiments on five widely used datasets with more than four practical metrics
    demonstrate Meta-SurDiff 
    establishing new state-of-the-art (SOTA) performance.
\end{itemize}

\section{Background}
\label{sec:background}

\subsection{Diffusion Probabilistic Model}
DDPM \citep{ho2020denoising} is a prominent example of probabilistic diffusion models \citep{song2020score,rombach2022high,shen2023non}, which comprises a forward diffusion process along with a reverse denoising process. 
Noise is incrementally introduced and ultimately converting the initial variable $\boldsymbol{y}_0$ into Gaussian noise $\boldsymbol{y}_T$ across $T$ steps:
\begin{align}
\begin{split}
    & q(\boldsymbol{y}_{1:T}|\boldsymbol{y}_0) = \prod_{t=1}^T q(\boldsymbol{y}_t|\boldsymbol{y}_{t-1}), \\
    & q(\boldsymbol{y}_t|\boldsymbol{y}_{t-1}) = \mathcal{N}(\sqrt{1-\beta_t} \boldsymbol{y}_{t-1}, \beta_t \textbf{I})
\end{split}
\end{align}
where $\beta_t$ is the noise level that typically set to a small constant. A notable characteristic of the forward process is that $q(\boldsymbol{y}_t|\boldsymbol{y}_0)=\mathcal{N}(\boldsymbol{y}_t; \sqrt{\alpha_t} \boldsymbol{y}_0, (1-\alpha_t) \textbf{I})$, $\alpha_t=\prod_{t=1}^T (1-\beta_t)$. Utilizing a Markov chain with trainable Gaussian transitions, the denoising process from $y_t$ back to $y_0$ unfolds as:
\begin{align}
\begin{split}
    & p_{\boldsymbol{\theta}}(\boldsymbol{y}_{0:T}) = p_{\boldsymbol{\theta}}(\boldsymbol{y}_T) \prod_{t=1}^T p_{\boldsymbol{\theta}}(\boldsymbol{y}_{t-1}|\boldsymbol{y}_t), \\
    & p_{\boldsymbol{\theta}}(\boldsymbol{y}_{t-1}|\boldsymbol{y}_t) = \mathcal{N}(\boldsymbol{\mu}_{\boldsymbol{\theta}}(\boldsymbol{y}_{t}, t), \sigma_{t}^2 \textbf{I})
\end{split}
\end{align}
where $\boldsymbol{\mu}_{\boldsymbol{\theta}}(\boldsymbol{y}_t, t)=\frac{1}{\sqrt{\alpha_t}} (\boldsymbol{y}_t - \frac{\beta_t}{\sqrt{1-\alpha_t}} \epsilon_{\boldsymbol{\theta}}(\boldsymbol{y}_t, t))$. Additionally, a noise prediction network $\epsilon_{\boldsymbol{\theta}}(\cdot)$ is adopted to minimize the regression loss $ \mathop{\min}_{\boldsymbol{\theta}} \mathbb{E}_{t,\boldsymbol{y}_0,\epsilon \sim \mathcal{N}(\textbf{0},\textbf{I})} \|\epsilon - \epsilon_{\boldsymbol{\theta}}(\boldsymbol{y}_t, t)\|_2^2$. 

\begin{figure*}[!t]
\begin{center}
\centerline{\includegraphics[width=0.8\textwidth]{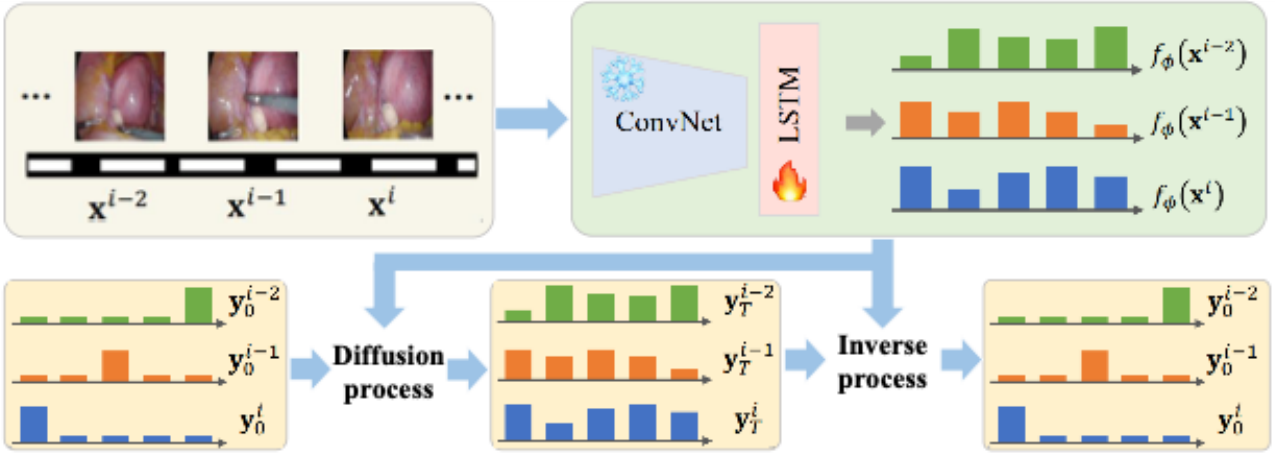}}
\caption{ Overview of Meta-SurDiff, it consists of a classification diffusion model and a re-weighting based meta-learning objective.
{\textbf{Top}}: We employ a simple yet effective backbone $f_{\phi}(\cdot)$, ConvNext $+$ LSTM, to capture coarse phase representations $f_{\phi}(\boldsymbol{x}^{i})$ for the $i$-th video frame, which serves as conditional inputs of the classification diffusion model. 
{\textbf{Bottom}}: the proposed classification diffusion model that utilizes $f_{\phi}(\boldsymbol{x}^{i})$ as prior and introduces covariate-dependence into both the forward and reverse diffusion processes aims to obtain precise frame-level phase distribution via reverse diffusion process given the coarse phase representations $f_{\phi}(\boldsymbol{x}^{i})$ (prior), as detailed in Sec. \ref{diffusion}. Taking unbalanced nature of surgical videos into consideration, we introduce a meta-learning objective to train the diffusion model via re-weighting the importance of each video frame, thus mitigating the risk of model being biased towards the majority surgical phases.
}
\label{fig:model}
\end{center}
\vspace{-4mm}
\end{figure*}

\subsection{Learning to Re-weight Examples 
}
\label{re-weight-bg}
Re-weighting the loss function is a widely employed tactic for addressing imbalanced data issue \citep{guo2022learning}. It treats the weight assigned to each instance as a trainable parameter, enabling the learning of a balanced model for both minority and majority categories through optimization of the weighted loss function. Typically, the optimal weight is optimized on a balanced meta dataset and can be expressed as:
\begin{align}
\begin{split}
    \boldsymbol{\theta}^*(\boldsymbol{w}) = \mathop{\arg\min}_{\boldsymbol{\theta}} \sum_{i=1}^N \boldsymbol{w}^i \mathcal{L}^i_{train}(\boldsymbol{\theta})
    \\ \boldsymbol{w}^* = \mathop{\arg\min}_{\boldsymbol{w}} \frac{1}{M} \sum_{j=1}^M \mathcal{L}_{meta}^j(\boldsymbol{\theta}^*(\boldsymbol{w}))
\end{split}
\end{align}
where $\boldsymbol{w} \in \mathbb{R}^{N}$ is weight vector
, $\boldsymbol{\theta}$ is classifier, $\mathcal{L}_{train}^i$ and $\mathcal{L}_{meta}^j$ are separately the loss functions 
on the unbalanced training dataset and the balanced meta dataset. 
The meta dataset is usually downsampled from the training dataset.

\section{Method}
\label{sec:method}

In this section, we introduce the meta-learning optimized classification diffusion model (Meta-SurDiff) for reliable Online Surgical Phase (OSP) recognition. The overview of our model is shown in Fig.\ref{fig:model}. We introduce the classification diffusion model in Sec. \ref{diffusion} and present the corresponding meta-learning optimization method in Sec. \ref{meta}.

\subsection{Classification Diffusion Model}
\label{diffusion}

\noindent \textbf{(1) Coarse surgical phase representations}:
Existing OSP recognition methods \citep{jin2021temporal,liu2023skit,ding2022exploring,zhao2022real,rivoir2024pitfalls} have primarily focused on learning strong spatial-temporal representations in surgical videos, which can last for several hours and exhibit strong dependencies among different phases. However, robustness of OSP recognition is rarely considered. In this paper, we advocate to adopt such strong phase representations as coarse conditional inputs, which are obviously point estimation, for the follow-up precise surgical phase distribution estimation. To reveal the superiority of our classification diffusion model in OSP recognition, inspired by \citep{rivoir2024pitfalls}, we use a simple yet effective backbone called ConvNext+LSTM to capture the coarse phase representations as follows:
\begin{align} \label{eq: condition}
\begin{split}
    \boldsymbol{z}^{i} = g_{\boldsymbol{\theta}}(f_{\boldsymbol{\phi}}(\boldsymbol{x}^i)), i=1,...,L
\end{split}
\end{align}
where $\boldsymbol{x}^i \in \mathbb{R}^{I}$ is the $i$-th video frame and there are totally $L$ frames, $f_{\boldsymbol{\phi}}(\cdot)$ is the backbone of ConvNext+LSTM, $g_{\boldsymbol{\theta}}(\cdot)$ denotes another learnable projector to align the dimension with the ground-truth label embedding $\boldsymbol{y}_0^i$ discussed later. Importantly, using LSTM for extracting spatial-temporal features is safe for OSP recognition because it avoids utilizing future frames during prediction. Additionally, we ensure the capability of the coarse phase representations by minimizing the cross entropy loss as $\mathcal{L}_{{CE}}=- \sum_{i=1}^{L}\sum_{c=1}^C \boldsymbol{y}^{i,c}_0 log \boldsymbol{z}^{i,c}$.

\noindent \textbf{(2) Forward diffusion process with coarse phase representations as priors}:
Unlike vanilla diffusion models that assume the endpoint of the diffusion process to be $\mathcal{N}(\textbf{0}, \textbf{I})$, 
we model the endpoint through the incorporation of the coarse phase representations $\boldsymbol{z}^i$ for $i=1,...,L$  which from HFNet as conditional inputs inspired by \citep{han2022card}, and we have $p(\boldsymbol{y}_T^i|\boldsymbol{z}^i)=\mathcal{N}(\boldsymbol{z}^i, \textbf{I})$. With a diffusion schedule $\beta_t \in (0,1)$ for $t=1,...,T$, the forward process is:
\begin{align} \label{eq:forward-1}
\begin{split}
    q(\boldsymbol{y}_t^i|\boldsymbol{y}_{t-1}^i, \boldsymbol{z}^i)=\mathcal{N}(\widetilde{\beta}_t \boldsymbol{y}_{t-1}^i + (1-\widetilde{\beta}_t) \boldsymbol{z}^i, \beta_t \textbf{I})
\end{split}
\end{align}
where $\widetilde{\beta}_t=\sqrt{1-\beta_t}$. Similar with DDPM \citep{ho2020denoising}, we can sample $\boldsymbol{y}_t^i$ given $\boldsymbol{y}_0^i$ with an arbitrary timestep $t$ as:
\begin{align} \label{eq:forward-2}
\begin{split}
    q(\boldsymbol{y}_t^i|\boldsymbol{y}_0^i,\boldsymbol{z}^{i})=\mathcal{N}(\sqrt{\alpha_t} \boldsymbol{y}_0^i + (1-\sqrt{\alpha_t}) \boldsymbol{z}^{i}, \widetilde{\alpha}_t \textbf{I})
\end{split}
\end{align}
where $\overline{\alpha}_t=1-\beta_t$, $\alpha_t=\prod_{t} \overline{\alpha}_t$, and $\widetilde{\alpha}_t=1-\alpha_t$. The mean value above is an interpolation between the ground-truth label embedding $\boldsymbol{y}_0^i$ and the conditional input $\boldsymbol{z}^{i}$ in Eq. \ref{eq: condition}. 

\noindent \textbf{(3) Reverse diffusion process with coarse phase representations as priors}: The primary objective of the reverse process is to progressively refine and sample denoised predictions from initially coarse estimates. This process involves complex mathematical transformations and probabilistic reasoning, aiming to reduce the uncertainty introduced during the forward process. Specifically, based on the detailed formulations in Eq. \ref{eq:forward-1} and \ref{eq:forward-2}, the posterior distribution of the forward process can be derived, as shown in Eq. \ref{eq:reverse-1}:
\begin{align} \label{eq:reverse-1}
    \begin{split}
    & q(\boldsymbol{y}^i_{t-1}|\boldsymbol{y}^i_{t},\boldsymbol{y}^i_0,\boldsymbol{z}^{i}) \propto q(\boldsymbol{y}^i_t|\boldsymbol{y}^i_{t-1},\boldsymbol{z}^{i}) q(\boldsymbol{y}^i_{t-1}|\boldsymbol{y}^i_0,\boldsymbol{z}^{i}) \\
    & \propto {\rm{exp}}\{-\frac{1}{2}[\frac{(\boldsymbol{y}^i_t-(1-\sqrt{\overline{\alpha}_t})\boldsymbol{z}^{i}-\sqrt{\overline{\alpha}_t}\boldsymbol{y}^i_{t-1})^2}{\beta_t} \\
    & + \frac{(\boldsymbol{y}^i_{t-1}-\sqrt{\alpha_{t-1}}\boldsymbol{y}^i_0-(1-\sqrt{\alpha_{t-1}})\boldsymbol{z}^{i})^2}{1-\alpha_{t-1}}]\} \\
    & \propto {\rm{exp}} \{-\frac{1}{2} [\boldsymbol{A} (\boldsymbol{y}^i_{t-1})^2-2 \boldsymbol{B} \boldsymbol{y}^i_{t-1}]\}
\end{split}
\end{align}
where $\boldsymbol{A} = \frac{1-\alpha_t}{\beta_t(1-\alpha_{t-1})}$ and $\boldsymbol{B} = \frac{\sqrt{\alpha_{t-1}}}{1-\alpha_{t-1}}\boldsymbol{y}^i_0+\frac{\sqrt{\overline{\alpha}_{t-1}}}{\beta_t} \boldsymbol{y}^i_t +(\frac{\sqrt{\overline{\alpha}_t}(\overline{\alpha}_t-1)}{\beta_t}+\frac{1-\sqrt{\alpha_{t-1}}}{1-\alpha_{t-1}}) \boldsymbol{z}^{i}$. Due to space limitations, we use $(\cdot)^2$ to replace $(\cdot)^T(\cdot)$ above, which does not affect the result of the derivation. According to properties of Gaussian distribution in Eq. 10.100 and Eq. 10.101 of \citep{bishop2006pattern}, the variance of posterior can be expressed as $\frac{1-\alpha_{t-1}}{1-\alpha_t} \beta_t$, and we have $\gamma_3=\frac{1-\alpha_{t-1}}{1-\alpha_t}$. The mean of posterior can be written as:
\begin{align} \label{eq:reverse-2}
\begin{split}
    &\widetilde{\mu}(\boldsymbol{y}^i_t,\boldsymbol{y}^i_0,\boldsymbol{z}^{i})=\frac{\beta_t \sqrt{\alpha_{t-1}}}{1-\alpha_t} \boldsymbol{y}^i_0 + \frac{1-\alpha_{t-1} \sqrt{\overline{\alpha}_t}}{1-\alpha_t} \boldsymbol{y}^i_t \\
    &+ (1+\frac{(\sqrt{\alpha_t-1})(\sqrt{\overline{\alpha}_t+\sqrt{\alpha_{t-1}}})}{1-\alpha_t})\boldsymbol{z}^{i}
\end{split}
\end{align}

For simplicity, we define $\widetilde{\boldsymbol{\mu}}=\gamma_0 \boldsymbol{y}^i_0 + \gamma_1 \boldsymbol{y}^i_t + \gamma_2 \boldsymbol{z}^{i}$ and we have:
\begin{align} \label{eq:gamma}
\begin{split}
    &\gamma_0=\frac{\beta_t \sqrt{\alpha_{t-1}}}{1-\alpha_t}, \quad \gamma_1=\frac{1-\alpha_{t-1} \sqrt{\overline{\alpha}_t}}{1-\alpha_t}, \\
    & \gamma_2=1+\frac{(\sqrt{\alpha_t-1})(\sqrt{\overline{\alpha}_t+\sqrt{\alpha_{t-1}}})}{1-\alpha_t}
\end{split}
\end{align}

Given a conditional input $z^{i}$, we can use the reverse diffusion process to generate precise phase distribution, which is expanded by multiple fine grained phase representations that closely resembles the ground-truth label embedding $\boldsymbol{y}^i_0$.


\begin{algorithm}[t]
\caption{Training of Meta-SurDiff}
\label{alg1}
\begin{algorithmic}
\State Initialize parameters $\boldsymbol{\Theta}$
\Repeat
    \State Draw mini-batch from $\mathcal{D}_{\text{train}}$ and $\mathcal{D}_{\text{meta}}$
    \State Draw $t \sim \mathrm{Uniform}(1, T)$
    \State Draw $\boldsymbol{\epsilon} \sim \mathcal{N}(\mathbf{0}, \mathbf{I})$
    \State Compute $\boldsymbol{z}^{\cdot}$ using Eq.~\ref{eq: condition}
    \State Compute loss per frame using Eq.~\ref{eq: intermediate}
    \State Update parameters using Eq.~\ref{eq: loss-1} and Eq.~\ref{eq: loss-2}
\Until{convergence}
\end{algorithmic}
\end{algorithm}

\begin{algorithm}[t]
\caption{Inference of Meta-SurDiff}
\label{alg2}
\begin{algorithmic}
\State Draw $\boldsymbol{x}^i$ from test dataset
\State Compute $\boldsymbol{z}^i$ using Eq.~\ref{eq: condition}
\For{$t = T$ to $1$}
    \State $\hat{\boldsymbol{y}}_0^i = \frac{1}{\alpha_t} \left( \boldsymbol{y}^i_t - (1 - \sqrt{\alpha_t}) \boldsymbol{z}^{i} - \sqrt{1 - \alpha_t} \, \epsilon_{\theta}(\boldsymbol{y}^i_t, \boldsymbol{z}^{i}, t) \right)$
    \If{$t \geq 1$}
        \State Draw $\boldsymbol{\epsilon} \sim \mathcal{N}(\mathbf{0}, \mathbf{I})$
        \State $\boldsymbol{y}^{i}_{t-1} = \gamma_0 \hat{\boldsymbol{y}}_0^i + \gamma_1 \boldsymbol{y}^i_t + \gamma_2 \boldsymbol{z}^i + \sqrt{\gamma_3 \beta_t} \, \boldsymbol{\epsilon}$
    \Else
        \State $\boldsymbol{y}^i_{t-1} = \hat{\boldsymbol{y}}_0^i$
    \EndIf
\EndFor
\end{algorithmic}
\end{algorithm}

\subsection{Re-weighting based Meta-Learning Objective}
\label{meta}
So far, we have improved the quality of phase representations preparing for reliable OSP recognition, however, unbalanced frame distribution among different phases still poses potential risk of overfitting. Therefore, we propose to use a re-weighting based meta-learning strategy to learn the parameters of our classification diffusion model.

According to Sec. \ref{re-weight-bg}, we first need to construct $\mathcal{L}_{train}$ and $\mathcal{L}_{meta}$, in practice, both of them should have the same formulation, taking $\mathcal{L}_{train}$ for example and we have:
\begin{align} \label{eq: metaloss}
\begin{split}
    \mathcal{L}_{train}&=\sum_{i=1}^L\mathbb{E}_{q(\boldsymbol{y}_{1:T}^i|\boldsymbol{y}_0^i,\boldsymbol{z}^i)}[log \frac{p_{\boldsymbol{\phi},\boldsymbol{\theta}}(\boldsymbol{y}_{0:T}^i|\boldsymbol{z}^i)}{q(\boldsymbol{y}_{1:T}^i|\boldsymbol{y}_0^i,\boldsymbol{z}^i)}] 
\end{split}
\end{align}
where the meaning of $\mathcal{L}_{train}$ is exactly maximizing the Evidence Lower BOund (ELBO) \citep{kingma2013auto} of $\sum_{i=1}^L log p_{\boldsymbol{\phi},\boldsymbol{\theta}}(\boldsymbol{y}_0^i|\boldsymbol{z}^i)$.
The specific form of Eq.\ref{eq: metaloss} can be divided into three items, for the $i$-th frame image from video, the items are separately $\mathbb{E}_q [-log p(\boldsymbol{y}_0^i|\boldsymbol{y}_1^i,\boldsymbol{z}^i)]$, $\mathbb{E}_q [KL(q(\boldsymbol{y}_T^i|\boldsymbol{y}_0^i,\boldsymbol{z}^i)||p(\boldsymbol{y}_T^i|\boldsymbol{z}^i))]$, and $\mathbb{E}_q[KL(q(\boldsymbol{y}_{t-1}^i|\boldsymbol{y}_t^i,\boldsymbol{y}_0^i,\boldsymbol{z}^i)||p_{\boldsymbol{\phi},\boldsymbol{\theta}}(\boldsymbol{y}_{t-1}^i|\boldsymbol{y}_t^i,\boldsymbol{z}^i))]$. 
As we can see, $\mathcal{L}_{train}$ guides the model to predict uncertainty while maintaining the capacity for accurate estimation of the fine-grained phase representation by the reversed diffusion. In practice, it is usually time-consuming to calculate the whole chain from the sampled timestep $t$ to $0$, therefore, we follow the reparameterization trick used in DDPM and construct $\epsilon_{\boldsymbol{\phi},\boldsymbol{\theta}}(\boldsymbol{y}_t^i,\boldsymbol{z}^i,t)$ to predict the forward diffusion noise $\boldsymbol{\epsilon}$ sampled from $\boldsymbol{y}_t^i$. The training objective $\mathcal{L}_{train}$ for the $i$-th frame can be carried out in a standard DDPM manner:
\begin{align}
\begin{split}
    \mathcal{L}_{\boldsymbol{\epsilon}^i}=\|\boldsymbol{\epsilon}^i-\epsilon_{\boldsymbol{\phi},\boldsymbol{\theta}}(\sqrt{\alpha_t} \boldsymbol{y}_0^i+\sqrt{\widetilde{\alpha}_t} \boldsymbol{\epsilon}+(1-\sqrt{\alpha_t})\boldsymbol{z}^i, \boldsymbol{z}^i,t)\|_2^2
\end{split}
\nonumber
\end{align}

Combining the noise prediction loss above and the objective in capturing meaningful coarse phase representations in Sec. \ref{diffusion} altogether, the intermediate objective is:
\begin{align} \label{eq: intermediate}
\begin{split}
    \mathcal{L}_{train}=\frac{\mathcal{L}_{\boldsymbol{\epsilon}} + \mathcal{L}_{CE}}{L}, \quad \mathcal{L}_{\boldsymbol{\epsilon}}=\sum_{i=1}^L \mathcal{L}_{\boldsymbol{\epsilon}^i}
\end{split}
\end{align}
Notably, we reuse the notation of $\mathcal{L}_{train}$ in Eq. \ref{eq: metaloss}, however, their meanings are different. The weight for each frame equals to $\frac{1}{L}$. In order to mitigate the negative impact of unbalanced frames across different phases on OSP recognition, we need to replace the equal weights with the dynamic weights. Drawn inspiration from \citep{guo2022learning, shu2019meta}, we adopt a re-weighting based meta-learning method to reassign the weights following the meta training and meta testing steps.

\noindent \textbf{(1) Meta training process:} 
We use a meta-weight net $h(\cdot, \boldsymbol{w})$ parameterized by $\boldsymbol{w}$, a two-layer MLP, to compute the frame-level weights. For convenience, we package $\boldsymbol{\phi}$ and $\boldsymbol{\theta}$ together and denote it as $\boldsymbol{\Theta}$.
We first update the meta-weight net in meta training process since the parameters $\boldsymbol{\Theta}$ of the classification diffusion model should be robust to unbalanced phase distribution, which heavily depends on the state of meta-weight net. Specifically, given $n$ and $m$ frames 
separately sampled from the training and meta datasets. Meta-weight net $h(\cdot, \boldsymbol{w})$ can be updated using:
\begin{align} \label{eq: loss-1}
\begin{split}
    & \boldsymbol{\hat{\Theta}}^t=\boldsymbol{\Theta}^t-\frac{\alpha}{n} \sum_{i=1}^n h(\mathcal{L}^i_{train};\boldsymbol{w}) \nabla_{\boldsymbol{\Theta}} \mathcal{L}_{train}^i \bigg|_{{\boldsymbol{\Theta}}^t} \\
    & \boldsymbol{w}^{t+1}=\boldsymbol{w}^t-\frac{\beta}{m} \sum_{i=1}^m \nabla_{\boldsymbol{w}} \mathcal{L}_{meta}^i \bigg|_{\boldsymbol{w}^t}
\end{split}
\end{align}
where $\alpha$ and $\beta$ now are the step sizes.

\noindent \textbf{(2) Meta testing process:} 
After obtaining the updated $\boldsymbol{w}^{t+1}$, the meta-weight net should be gradually capable of reassigning proper weights for unbalanced video frames.
Consequently,
we use the updated $\boldsymbol{w}^{t+1}$ to update the parameters in $\boldsymbol{\Theta}$ of our model, which can be expressed as:
\begin{align} \label{eq: loss-2}
\begin{split}
    \boldsymbol{\Theta}^{t+1}=\boldsymbol{\Theta}^t-\frac{\alpha}{n} \sum_{i=1}^n h(\mathcal{L}_{train}^i;\boldsymbol{w}^{t+1}) \nabla_{\boldsymbol{\Theta}} \mathcal{L}_{train}^i \bigg|_{\boldsymbol{\Theta}^t}
\end{split}
\end{align}
The Meta-SurDiff is trained through updating parameters iteratively between the two meta-learning processes on diverse mini-batch of video frames. 
Right now, Meta-SurDiff is properly handling the uncertainties of video quality and unbalanced distribution,
thereby is beneficial for facilitating reliable OSP recognition. The pseudocodes of training and inference are separately presented in Alg. \ref{alg1} and Alg. \ref{alg2}.

\section{Experiment}
\label{sec:experiment}

\subsection{Experimental Setup}
\label{setup}

\textbf{Datasets:}
Five surgical phase recognition datasets are utilized to extensively evaluate our model, including Cholec80\citep{twinanda2016endonet}, M2Cai16\citep{twinanda2016miccai}, AutoLaparo\citep{wang2022autolaparo}, OphNet\citep{hu2024ophnetlargescalevideobenchmark}, and NurViD \citep{hu2023nurvidlargeexpertlevelvideo}. Table \ref{tab-data} is basic statistical information of these datasets, and details are depicted in Appendix \ref{details-data}. Cholec80, M2Cai16, and AutoLaparo are datasets of laparoscopic surgery, OphNet belongs to dataset of ophthalmic surgery, and NurViD is dataset of daily care.

\begin{table}[h]
\caption{Statistics of datasets. TR/VAL/TE No. is the number of training, validation, and testing videos. C No. is the phase number.}
\vspace{-2mm}
\label{tab-data}
\vskip 0.05in
\begin{center}
\begin{small}
\begin{sc}
\resizebox{0.75\textwidth}{!}{%
\begin{tabular}{c|c|c|c|c}
\toprule
Dataset &Duration &fps(f/s) &Tr/Val/Te No. & C No. \\
\midrule
Cholec80 & 38${\rm{min}}$26${\rm{s}}$ &25 &40/-/40 &7 \\
  MeCai16 &38${\rm{min}}$25${\rm{s}}$ &25 &27/-/14 &8 \\
  AutoLaparo &66${\rm{min}}$07${\rm{s}}$ &25 &10/4/7 &7 \\
  OphNet & 5${\rm{min}}$37${\rm{s}}$ & 25 & 445/144/154 & 96 \\
  NurViD & 32${\rm{s}}$ & - &  3906/587/1122 &  177 \\
\bottomrule
\end{tabular}%
}
\end{sc}
\end{small}
\end{center}
\vspace{-2mm}
\end{table}

\noindent \textbf{Evaluation metrics:}
We use four widely used metrics including accuracy (Acc), precision (Pr), recall (Re), and Jaccard (Ja) to evaluate the OSP recognition performance. Additionally, since prior surgical phase recognition methods do not explicitly model the uncertainty in surgical videos, we leverage Prediction Interval Width (PIW) and Paired Two Samples t-Test (PTST) to quantify the model's uncertainty. Please refer to Appendix \ref{piw-ttest} for more details.
Due to the subjective nature of manual labeling in surgical videos and the ambiguous boundaries between adjacent surgical stages which are noted by \citep{yi2022not}, Cholec80 and M2Cai16 datasets adopt lenient  boundary metrics to access model performance. Specifically, frames predicted belonging to adjacent stages within a 10 seconds window before and after a phase transition are also deemed correct. As for OphNet and Nurvid, we implement comparisons following the task settings and evaluation metrics outlined in their paper 
.

\noindent \textbf{Baselines:}
We compare our Meta-SurDiff with some most recently proposed competitive baseline methods, such as PitBN \citep{rivoir2024pitfalls}, SKiT\citep{liu2023skit}, CMTNet \citep{yue2023cascade}, LAST \citep{tao2023last}, TMRNet \citep{jin2021temporal}, Trans-SVNet \citep{gao2021trans}, TeCNO \citep{czempiel2020tecno}, SV-RCNet \citep{jin2017sv} and so on.
The results are reported from their original papers or reproduced using their available official codes.

\begin{table}[t]
\caption{The results ($\%$) of Meta-SurDiff V.S. other competitors on Cholec80 dataset. The best results are marked in bold.}
\vspace{-2mm}
\label{tab-c80}
\begin{center}
\begin{small}
\begin{sc}
\resizebox{0.75\textwidth}{!}{%
\begin{tabular}{c|ccccc}
\toprule
    \texttt{Methods}
    &\texttt{R} & \texttt{Acc} &\texttt{Pr} &\texttt{Re} &\texttt{Ja} \\
\midrule
Trans-SVNet\citep{gao2021trans}   &\checkmark  &$90.3 \pm 7.1$&90.7      &88.8   & 79.3 \\ 
TeSTra\citep{zhao2022real} &  &$90.1 \pm 6.6$  &82.8      &83.8   & 71.6 \\
Dual Pyramid\citep{chenDP2022} &  &91.4  &85.4      &86.3   & 75.4 \\
OperA\citep{Czempiel_2021} &\checkmark  &$90.2 \pm 6.1$  &84.2      &85.5   & 73.0 \\
CMTNet\citep{yue2023cascade}   &\checkmark  &$92.9 \pm 5.9 $ &90.1      &92.0   & 81.5 \\
LAST\citep{tao2023last}   &\checkmark  &$93.1 \pm 4.7 $ &89.3&90.1   & 81.1 \\
LoViT\citep{liu2023lovit}   &\checkmark  &$92.4 \pm 6.3 $ &89.9      &90.6   & 81.2 \\
SKiT\citep{liu2023skit} &\checkmark  &$93.4 \pm 5.2 $ &90.9      &91.8   & 82.6 \\
PitBN\citep{rivoir2024pitfalls}   &\checkmark  &$93.5 \pm 6.5 $ &90.0      &91.9   & 82.9 \\
SurgPLAN++\citep{chen2025surgplanuniversalsurgicalphase} &  & 92.7 & 91.1  & 89.8  & 81.4 \\
SR-Mamba\citep{cao2024srmambaeffectivesurgicalphase} & \textbf{$\checkmark$} & 92.6 & 90.3 & 90.6 & 81.5 \\
SPR-Mamba\citep{zhang2024sprmambasurgicalphaserecognition} & \textbf{$\checkmark$} & 93.1 & 89.3 & 90.1 & 81.4  \\
Meta-SurDiff (Ours)   &  &$94.2 \pm 4.3 $ &89.6      &90.0   & 81.7 \\
Meta-SurDiff (Ours)   &\checkmark  &$\textbf{95.3} \pm \textbf{4.1} $ & \textbf{92.9}      &\textbf{93.1}   & \textbf{86.0} \\
\bottomrule
\end{tabular}%
}
\end{sc}
\end{small}
\end{center}
\vspace{-4mm}
\end{table}

\begin{table}[t]
\caption{The results ($\%$) of Meta-SurDiff V.S. other competitors on AutoLaparo dataset. The best results are marked in bold.}
\vspace{-2mm}
\label{tab-auto}
\begin{center}
\begin{small}
\begin{sc}
\resizebox{0.75\textwidth}{!}{%
\begin{tabular}{c|ccccc}
\toprule
    \texttt{Methods}
    &\texttt{R} & \texttt{Acc} &\texttt{Pr} &\texttt{Re} &\texttt{Ja} \\
\midrule
SV-RCNet\citep{2018SV-RCNETJin}   &  &75.6  &64.0      &59.7   & 47.2 \\
TeCNO\citep{czempiel2020tecno} &  &77.3  &66.9      &64.6   & 50.7 \\
TMRNet\citep{jin2021temporal}   &  &78.2  &66.0      &61.5   & 49.6 \\
Trans-SVNet\citep{gao2021trans}   &  &78.3  &64.2      &62.1   & 50.7 \\ 
LoViT\citep{liu2023lovit}   &  &$81.4\pm7.6 $ & \textbf{85.1}      &65.9   & 56.0 \\
SKiT\citep{liu2023skit} &  &$82.9 \pm 6.8 $ &81.8      &70.1   & 59.9 \\
PitBN\citep{rivoir2024pitfalls}   &  &$83.7 \pm 6.6$  &79.5      &67.7   & 58.8 \\
Meta-SurDiff (Ours)   &  &$\textbf{85.8} \pm \textbf{6.0} $ &82.3      &\textbf{71.1}   & \textbf{61.2} \\
\bottomrule
\end{tabular}%
}
\end{sc}
\end{small}
\end{center}
\vspace{-4mm}
\end{table}

\noindent \textbf{Implementation details:}
We utilize ConvNeXt \citep{liu2022convnet} pretrained on ImageNet-1K \citep{krizhevsky2017imagenet} to extract spatial features from videos, followed by LSTM for temporal feature fusion. During training, we freeze the earlier blocks of ConvNeXt and only update the parameters of its last block. 
To generate meaningful conditional inputs for learning our classification diffusion model, we initially pretrain the backbone of ConvNeXt $+$ LSTM, using standard cross-entropy loss on unbalanced training datasets.
We employ AdamW \citep{kingma2014adam} to optimize the model, with separate learning rates of 1e-5 for $\boldsymbol{\Theta}$ and 1e-3 for $\boldsymbol{w}$, without weight decay. To ensure fair comparisons, we maintain batch size of 1 and the time window length of 256, consistent with other competitors. All experiments are conducted on a single NVIDIA A100 80GB PCIe GPU. More details can be found in Appendix \ref{archi}.

\subsection{Main Results}
\subsubsection{Quantitative Results and Analysis}
\label{sec4-2-1}

\noindent \textbf{(1) Online surgical phase recognition:}
We conduct comprehensive studies comparing Meta-SurDiff with other SOTA methods for OSP recognition on Cholec80, AutoLaparo, M2Cai16, OphNet, and NurViD datasets. Quantitative results for these datasets are separately reported in Table \ref{tab-c80}, Table \ref{tab-auto}, Table \ref{tab-m2}, Table \ref{tab-ophnet}, and Table \ref{tab-Nur}. Meta-SurDiff significantly outperforms most of the competitors, such as SKiT and LAST, across various metrics including accuracy (Acc), precision (Pr), recall (Re), and Jaccard (Ja). For example, Meta-SurDiff shows improvements on Cholec80 with increase of 1.8$\%$ in \texttt{Acc}, $1.8\%$ in \texttt{Pr}, 1.1$\%$ in \texttt{Re}, and 3.1$\%$ in \texttt{Ja} compared to the second-best method. Additionally, Meta-SurDiff delivers superior results in \texttt{Pr}, \texttt{Re}, and \texttt{Ja}, due to effectively addressing unbalanced effects. 
Meta-SurDiff also achieves lower standard deviations of \texttt{Acc}, with reductions of 0.6$\%$, 0.8$\%$, and 0.5$\%$ on Cholec80, AutoLaparo, and M2Cai16 datasets, respectively, compared to the second-best method. 
We attribute these notable improvements to effectiveness of Meta-SurDiff in addressing the surgical video issues of frame ambiguity and unbalanced phase distribution.

\noindent \textbf{(2) Complexity analysis:}
We report running time, CPU and GPU memories at training time in Table \ref{tab-c80-train} on Cholec80 dataset. At test time, we select the diffusion timestep to be $T=1000$. To accelerate prediction speed, we employ the DDIM \citep{song2020denoising} sampling strategy, reducing the total sampling requirement effectively to $T=100$. On one hand, we conduct comparative experiments using different diffusion timesteps and depict results in Table \ref{tab-c80-infer}. 
On the other hand, we also compare the complexity of Meta-SurDiff with other competitors.
Overall, our model achieves a satisfactory balance between performance and real-time efficiency.

\begin{table}[t]
\caption{The results ($\%$) of Meta-SurDiff V.S. other competitors on M2Cai16 dataset. The best results are marked in bold.}
\vspace{-2mm}
\label{tab-m2}
\begin{center}
\begin{small}
\begin{sc}
\resizebox{0.75\textwidth}{!}{%
\begin{tabular}{c|ccccc}
\toprule
    \texttt{Methods}
    &\texttt{R} & \texttt{Acc} &\texttt{Pr} &\texttt{Re} &\texttt{Ja} \\
\midrule
SV-RCNet\citep{2018SV-RCNETJin}   &\checkmark  & $81.7 \pm 8.1 $ & 81.0      &81.6   & 65.4 \\
OHFM\citep{2019OHFMYi}   &\checkmark  & $85.2 \pm 7.5$  &--      &--   & 68.8 \\
TMRNet\citep{jin2021temporal}   &\checkmark  &$87.0 \pm 8.6$  &87.8      &88.4   & 75.1 \\
Not-E2E\citep{yi2022not} &\checkmark  &$84.1 \pm 9.6 $ &--      &88.3   & 69.8 \\
Trans-SVNet\citep{gao2021trans}   &\checkmark  &$87.2 \pm 9.3$  &88.0      &87.5   & 74.7 \\
CMTNet\citep{yue2023cascade}   &\checkmark  &$88.2 \pm 9.2$  &88.3      &88.7   & 76.1 \\
LAST\citep{tao2023last}   &\checkmark  &$91.5 \pm 5.6$  &86.3      &88.7   & 77.8 \\
PitBN\citep{rivoir2024pitfalls}   &\checkmark  &$91.1 \pm 7.2$  &90.0      &92.5   & 81.4 \\
Meta-SurDiff (Ours)   &\checkmark  &$\textbf{92.2} \pm \textbf{5.1}$  &\textbf{91.5}      &\textbf{92.7}   & \textbf{82.9}  \\
\bottomrule
\end{tabular}
}
\end{sc}
\end{small}
\end{center}
\vspace{-4mm}
\end{table}

\begin{table}[t]
\caption{Results on the OphNet dataset using metrics following OphNet. The best results are marked in bold.}
\vspace{-2mm}
\label{tab-ophnet}
\begin{center}
\begin{small}
\begin{sc}
\resizebox{0.70\textwidth}{!}{%
\begin{tabular}{c c c c c}
\toprule
    \multicolumn{1}{c}{Method} & \multicolumn{1}{c}{Acc-top1} & \multicolumn{1}{c}{Acc-top5} &
        \multicolumn{1}{c}{Params} &
        \multicolumn{1}{c}{Time} \\
        \midrule
         X-CLIP16\citep{ni2022expandinglanguageimagepretrainedmodels} & 64.8 & 89.3 & 194.9M & 216ms \\
         X-CLIP32\citep{ni2022expandinglanguageimagepretrainedmodels} & 71.2 & 91.6 & 194.9M & 243ms\\
         Ours & 69.7 & 91.4 & \textbf{32.8}M & \textbf{170}ms \\
         Ours+X-CLIP32 & \textbf{73.0} & \textbf{92.4} & 194.9M & 256ms \\
\bottomrule
\end{tabular}%
}
\end{sc}
\end{small}
\end{center}
\vskip -0.15in
\end{table}

\begin{table}[t]
\caption{The results ($\%$) of Meta-SurDiff V.S. other competitors on NurViD. The best results are marked in bold.}
\label{tab-Nur}
\vspace{-2mm}
\begin{center}
\begin{small}
\begin{sc}
\resizebox{0.75\textwidth}{!}{%
\begin{tabular}{c cccc}
\toprule
   Method & Many(9) & Medium(66) & Few(87) & All(162)  \\
            \midrule
             SlowFast\citep{feichtenhofer2019slowfastnetworksvideorecognition} & 29.8 & 15.5  & 7.9  & 21.1 \\
             C3D\citep{tran2015learningspatiotemporalfeatures3d} & 28.1 & 14.6 & 7.3 & 22.8 \\
             I3D\citep{carreira2018quovadisactionrecognition} &  31.3 & 14.8 & 8.2 & 21.5 \\
             Ours & \textbf{34.2} & \textbf{22.5} & \textbf{19.3} & \textbf{26.0}  \\
\bottomrule
\end{tabular}%
}
\end{sc}
\end{small}
\end{center}
\vskip -0.15in
\end{table}

\begin{table}[t]
\caption{Ablative results of Meta-SurDiff on the Cholec80 dataset.}
\vspace{-2mm}
\label{tab-ablation}
\begin{center}
\begin{small}
\begin{sc}
\resizebox{0.55\textwidth}{!}{%
\begin{tabular}{c|c|c|c|c|c}
\toprule
    CDM & MLO &Acc &Pr &Re &Ja \\ \midrule
  & &$93.5 \pm 6.5$ &90.0 &91.9 &82.9 \\ 
  \checkmark & &$94.2 \pm 5.2$ &91.1 &92.1 &83.4 \\ 
  &\checkmark &$94.5 \pm 4.8$ &92.3 &92.7 &85.2 \\
  \checkmark &\checkmark &$\textbf{95.3} \pm 4.1$ & \textbf{92.9}  & \textbf{93.1 }& \textbf{86.0} \\
\bottomrule
\end{tabular}%
}
\end{sc}
\end{small}
\end{center}
\vskip -0.15in
\end{table}

\begin{table}[t]
\caption{Ablation studies on Train Time, CPU and GPU Memories.}
\label{tab-c80-train}
\vspace{-2mm}
\begin{center}
\begin{small}
\begin{sc}
\resizebox{0.60\textwidth}{!}{%
\begin{tabular}{c c ccc}
\toprule
  \multicolumn{1}{c}{MLO} & \multicolumn{1}{c}{CDM} & \multicolumn{1}{c}{Train Time} & \multicolumn{1}{c}{CPU Mem} & \multicolumn{1}{c}{GPU Mem} \\
            \midrule
            \textbf{\(\times\)} & \textbf{\(\times\)} & 19:34:52 & 4.93G & 7.99G \\
            \textbf{\(\checkmark\)} & \textbf{\(\times\)} &  30:12:21 & 6.72G & 9.44G \\
            \textbf{\(\times\)} & \textbf{\(\checkmark\)} & 25:58:19 & 8.18G & 9.37G \\
            \textbf{\(\checkmark\)} & \textbf{\(\checkmark\)} & 34:35:16 & 10.22G & 10.82G \\
\bottomrule
\end{tabular}%
}
\end{sc}
\end{small}
\end{center}
\vskip -0.15in
\end{table}

\begin{table}[t]
\caption{Complexity and running time analysis on Cholec80 dataset.}
\vspace{-2mm}
\label{tab-c80-infer}
\begin{center}
\begin{small}
\begin{sc}
\resizebox{0.8\textwidth}{!}{%
\begin{tabular}{c|c|c|c|c}
\toprule
    Methods &Params (M) &time (ms) & GFLOPs &Acc (\%) \\ \midrule
  TeCNO\citep{czempiel2020tecno} & 24.69 &19  &4.11 &$88.6 \pm 7.8$ \\
  TMRNet\citep{jin2021temporal} & 63.02 &26  &8.29 &$90.1 \pm 7.6$ \\
  Trans-SVNet\citep{gao2021trans} & 24.72 &19  &4.15 &$90.3 \pm 7.1 $\\
  Not-E2E\citep{yi2022not} & 22.73 &49  &5.72 &$91.5 \pm 7.1$ \\
  CMTNet\citep{yue2023cascade} & 26.63 &33  &5.56 &$92.9 \pm 5.9 $\\
  LAST\citep{tao2023last} & 117.26 &86  &15.49  &$93.1 \pm 4.7 $\\
  \midrule
  Meta-SurDiff-10 & 21.44 &9 &5.81  &$95.0 \pm 4.3$ \\
  Meta-SurDiff-100 & 21.44 &76  &11.96 &$95.3 \pm 4.1 $\\
  Meta-SurDiff-500 & 21.44 &367  &39.27 &$95.5 \pm 4.1$ \\
\bottomrule
\end{tabular}%
}
\end{sc}
\end{small}
\end{center}
\vskip -0.15in
\vspace{-2mm}
\end{table}

\noindent \textbf{(3) Ablation study:} We verify the effects of different components on our proposed Meta-SurDiff and report results in Table \ref{tab-ablation}. \texttt{CDM} represents whether using Classification Diffusion Model (CDM) to process the coarse phase representations, also known as conditional inputs. \texttt{MLO} refers to whether employing Meta-Learning Objective (MLO) to re-weight the intermediate loss function in Eq. \ref{eq: intermediate}.
On one hand, either equipping \texttt{CDM} or \texttt{MLO} can consistently improve recognition performances across all metrics. On the other hand, combining both \texttt{CDM} and \texttt{MLO} altogether can further boost recognition. Additionally, we are surprised to observe that equipping \texttt{CDM} can enhance metrics like \texttt{Pr},\texttt{Re},\texttt{Ja} that reflect unbalance distribution. It implies that \texttt{CDM} has capability to ameliorate robustness on unbalanced surgical videos thanks to its precise frame-level distribution modeling.
Moreover, the ablative results on the occupation of training resources can be further found in Table \ref{tab-c80-train}.

\noindent \textbf{(4) Hyper-parameters analysis:}
We analyze the impacts of four hyper-parameters closely related with our Meta-SurDiff, including fine-tuning scope, meta-dataset size, backbone selection, and surgical videos' background.

\noindent \textbf{i) The impact of fine-tuning scope.} As mentioned in the implementation details of Sec. \ref{setup} that we initially pre-train Meta-SurDiff using standard cross-entropy loss function to obtain meaningful conditional embeddings. Therefore, we investigate the performances V.S. the scope of optimized parameters during fine-tuning and report the results in Table \ref{tab-scope}, where \texttt{C} represents we only fine-tune the classification diffusion model (CDM) parameters. \texttt{LSTM+C} refers to that we fine-tune parameters of both the LSTM and CDM. \texttt{ConvNeXt$^\#$+LSTM+C} denotes that we fine-tune the parameters of the last block in ConvNeXt, LSTM, and CDM. \texttt{ConvNeXt+LSTM+C} is that we fine-tune the whole parameters of ConvNeXt, LSTM, and CDM. We find that fine-tuning parameters with proper scope is beneficial for accelerating performances, which might be able to attribute to the fact that basic and general representations of ConvNeXt is essential for reliable OSP recognition of Meta-SurDiff.

\begin{figure*}[t]
    \centering
    \begin{minipage}[t]{0.53\textwidth}
    \vspace{0pt} 
        \centering        \includegraphics[width=0.9\linewidth]{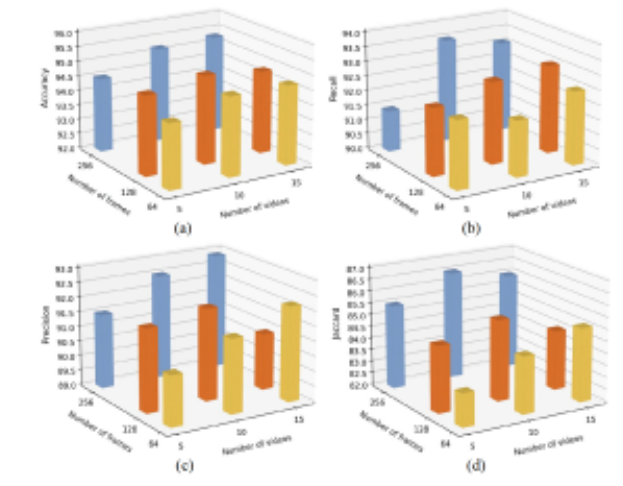}
        \captionsetup{width=\linewidth}
        \caption{Results change with the size of meta dataset on Cholec80 dataset.}
        \label{fig:hyper}
    \end{minipage}
    \hfill
    \begin{minipage}[t]{0.43\textwidth}
    \vspace{0pt} 
        \centering        \includegraphics[width=0.9\linewidth]{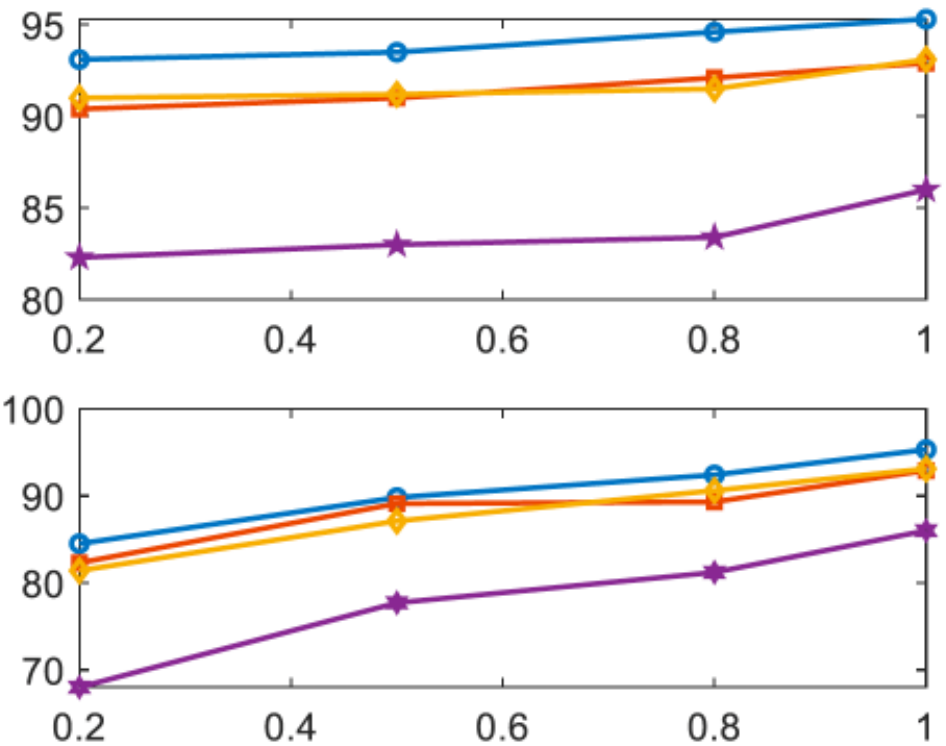}
        \captionsetup{width=\linewidth}
        \caption{Performance changes with the number of training frames in Cholec80 dataset. \textbf{Top:} Results of ignoring training labels. \textbf{Bottom:} Results of ignoring training frames.}
        \label{fig:lowdata}
    \end{minipage}
\end{figure*}

\noindent \textbf{ii) The impact of meta-dataset size.}
We study performances V.S. the scale of meta dataset and report results in Fig. \ref{fig:hyper}. We observe that our model achieves consistent recognition performances even with a small scale meta dataset, demonstrating its significance and suitability for practical online surgical phase recognition applications.

\noindent \textbf{iii) The impact of backbone selection.}
To verify that our model can achieve consistent improvement among various backbones, we conduct experiments on Cholec80 dataset and report results in Table \ref{tab-c80-bb}. 
Additionally, Table \ref{tab-ophnet} presents the results of Meta-SurDiff V.S. different backbones on OphNet dataset. As we can see taht by applying Meta-SurDiff to the X-CLIP model improves 
ACC-Top1 and ACC-Top5 by $0.9\%$ and $0.8\%$, respectively, further demonstrating that our model can be seamlessly compatible with existing backbones.

\noindent \textbf{iv) The impact of surgical videos' background.}
We conduct experiments on OphNet dataset to study the impact of background on OSP recognition since OphNet dataset is partially annotated, and report the results in Table. \ref{tab-Oph-bg}. In \texttt{IGNORE BG}, we directly ignore the unlabeled frames while training the model. In \texttt{IGNORE BGL}, we leverage the visual information of the unlabeled frames to train the model. In \texttt{USE 1 BGL}, we set the unlabeled frames with a same and fixed pseudo label when we train the model. According to the results, we find that \texttt{IGNORE BGL} achieves the best performance since it effectively utilizes the unsupervised spatial-temporal information to learn the model. Notably, \texttt{USE 1 BGL} wins the last place. And we attribute this to the fact that given the same pseudo label may mislead the potential disciminative representations in backgrounds.


\noindent \textbf{(5) OSP recognition in low-data regime:}
In practice, the annotation of surgical videos is laborious and time consuming, therefore, verifying the effectiveness of our model under low data regime is crucial. We conduct experiments using Cholec80 under two data-limited scenarios, and report the results in Fig. \ref{fig:lowdata}. 
Meta-SurDiff could maintain robust OSP recognition results even when training frames are limited. Taking Fig. \ref{fig:lowdata} \textbf{Top} for example, we randomly ignore labels of $\gamma \%$ ($\gamma \in [1, 0]$) training frames along with other fully supervised frames to learn the model. That is, we use the spatial-temporal power of LSTM to effectively capture meaningful phase representations, which should be beneficial for our classification diffusion model training and inference even supervised information is not fully explored.

\begin{table}[t]
\caption{Results on the OphNet dataset regarding the effect of background frames using X-CLIP32 backbone in \citep{hu2024ophnetlargescalevideobenchmark}.}
\label{tab-Oph-bg}
\vspace{-2mm}
\begin{center}
\begin{small}
\begin{sc}
\resizebox{0.60\textwidth}{!}{%
\begin{tabular}{c cccc}
\toprule
  \multicolumn{1}{c}{Method} & \multicolumn{1}{c}{Acc} & \multicolumn{1}{c}{Precision} & \multicolumn{1}{c}{Recall} & \multicolumn{1}{c}{Jaccard} \\
        \midrule
          ignore bg & 73.0 & 63.2 & 56.3 & 55.6 \\
          \textcolor{black}{ignore bgl} & 76.3 & 64.3 & 58.4 & 59.3 \\
          use 1 bgl & 70.0 & 56.2 & 51.7 & 52.5 \\
\bottomrule
\end{tabular}%
}
\end{sc}
\end{small}
\end{center}
\vskip -0.15in
\end{table}

\begin{table}[t]
\caption{Results on the Cholec80 dataset using different backbones. The best results are marked in bold.}
\vspace{-2mm}
\label{tab-c80-bb}
\begin{center}
\begin{small}
\begin{sc}
\resizebox{0.75\textwidth}{!}{%
\begin{tabular}{c|ccccc}
\toprule
    \texttt{Methods}
    &\texttt{R} & \texttt{Acc} &\texttt{Pr} &\texttt{Re} &\texttt{Ja} \\
\midrule
Meta-SurDiff+ResNet & \textbf{$\checkmark$} & 94.8 & 90.7 & 92.8 & 84.5 \\
Meta-SurDiff+ViT & \textbf{$\checkmark$} & 95.0 & 91.7 & \textbf{93.2} & 85.4 \\
Meta-SurDiff+ConvNext(Ours)   &  &$94.2 \pm 4.3 $ &89.6      &90.0   & 81.7 \\
Meta-SurDiff+ConvNext(Ours)   &\checkmark  &$\textbf{95.3} \pm \textbf{4.1}$  & \textbf{92.9}     &93.1 & \textbf{86.0} \\
\bottomrule
\end{tabular}%
}
\end{sc}
\end{small}
\end{center}
\vskip -0.15in
\end{table}

\begin{table}[t]
\caption{The results ($\%$) on the scope of optimized parameters under Cholec80 dataset.}
\vspace{-2mm}
\label{tab-scope}
\begin{center}
\begin{small}
\begin{sc}
\resizebox{0.60\textwidth}{!}{%
\begin{tabular}{c|c|c|c|c}
\toprule
  Range &Acc &Pr &Re &Ja \\ \midrule
  C &$93.4 \pm 5.2 $&90.7 &90.7 &82.7 \\ 
  LSTM+C &$94.2 \pm 4.5 $&92.0 &91.0 &83.0 \\ 
  ConvNeXt$^\#$+LSTM+C &$95.3 \pm 4.1$ &92.9 &93.1 &86.0 \\
  ConvNeXt+LSTM+C &$90.6 \pm 5.6$ &88.7 &89.6 &81.9 \\
\bottomrule
\end{tabular}%
}
\end{sc}
\end{small}
\end{center}
\vspace{-4mm}
\end{table}

\begin{table}[t]
    \centering
    \caption{PIW ($\times$ 100) and t-test on Cholec80 dataset.}
    \vspace{2mm}
    \resizebox{0.75\textwidth}{!}{
    \begin{tabular}{cccccc}
    \toprule
        \multirow{2}*{Class} & \multirow{2}*{Accuracy} & \multicolumn{2}{c}{PIW} & \multicolumn{2}{c}{Acc by t-Test} \\ 
        \cmidrule{3-6}
        & & Correct & Incorrect & Reject & Not-Reject (count) \\
    \midrule
        all & 81.3\% & 0.65 & 13.40 & 91.2\% & 50.8\%(134) \\ 
        0 & 39.0\% & 0.43 & 1.45 & 39.0\% & 20.0\%(5) \\
        1 & 97.4\% & 0.13 & 10.69 & 97.4\% & 60.0\%(10) \\
        2 & 83.4\% & 0.51 & 8.12 & 83.5\% & 37.5\%(8) \\
        3 & 96.5\% & 0.39 & 17.10 & 96.5\% & 52.4\%(21) \\
        4 & 88.8\% & 0.97 & 27.31 & 88.9\% & 50.0\%(8) \\
        5 & 78.9\% & 3.07 & 23.99 & 79.2\% & 43.8\%(48) \\
        6 & 84.6\% & 3.24 & 45.89 & 84.8\% & 64.7\%(34) \\ 
    \bottomrule
    \end{tabular}}
    \label{tab-uncertain}
    \vspace{-2mm}
\end{table}

\noindent \textbf{(6) Uncertainty estimation:}
We present the results of our Meta-SurDiff on uncertainty estimation to evaluate the frame-level recognition confidence under the scope of the entire Cholec80 test dataset in Table \ref{tab-uncertain}. Since prior surgical phase recognition methods do not explicitly model the uncertainty in surgical videos, we use PIW and PTST to quantify the model’s uncertainty. Specifically, for each test frame, we generate 100 predictions through the reverse diffusion process, resulting in a $100 \times 7$ matrix. We then compute PIW and PTST based on this matrix.
After obtaining the PIW and the PTST from each test frame, we divide the test dataset into two groups by the correctness of majority-vote predictions. We calculate the average PIW of the true phase within each group. We split the test instances by t-test rejection status, and compute the mean accuracy in each group. More details can be found in Appendix \ref{piw-ttest}.

As we can see that the mean PIW of the ground truth label among the correct predictions is (10$\times$) narrower than that of the incorrect predictions, indicating that Meta-SurDiff can make correct predictions with much smaller variations. Furthermore, when comparing the mean PIWs across different phases, we observe that the phase indexed as $0$ has the lowest accuracy at $39.0\%$ and its incorrect prediction interval is much smaller than other phases. All these evidences suggest that the uncertainties of phase $0$ could be especially significant.
Moreover, we observe that the accuracy of test instances rejected by the t-Test is significantly higher than that of the not-rejected ones, both across the entire test dataset and within each phase. We point out that these metrics reflect confidence of Meta-SurDiff in the correctness of predictions and have the potential to be applied in mitigating risks during surgical evaluation.
Such uncertainty estimation can be used to decide whether to accept the recognition results or to refer the instance to experts for further evaluation.

\begin{figure*}[!t]
\centering
\centerline{\includegraphics[width=0.95\linewidth]{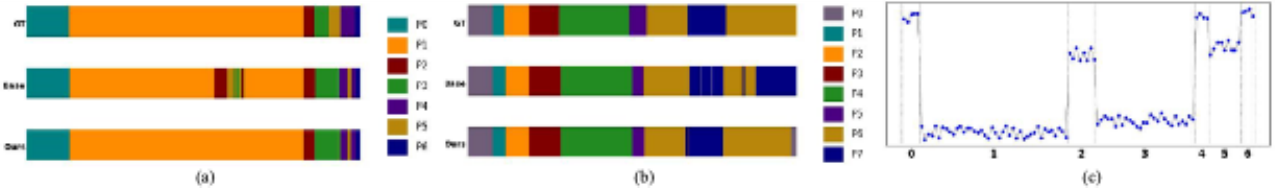}}
\vspace{-2mm}
\caption{(a) and (b) are ribbon diagrams of ground truth labels, baseline method, and our proposed Meta-SurDiff from the top to the bottom under Cholec80 and M2Cai16 datasets. (c) The learned weight vectors on Cholec80 dataset, where x-axis is the frames from the current mini-batch and we only mark their labels for clarity. For example, frames from the 1st phase is much more than those from the 0-th phase, obviously, the weights for the 0-th frames are higher than those for the 1-st frames.
}
\label{fig:vis}
\end{figure*}

\subsubsection{Qualitative Results and Analysis}

\textbf{(1) The predicted ribbon charts.} To intuitively reveal the effectiveness of our model in handling unbalanced phase distribution, We employ ConvNeXt $+$ LSTM optimized with standard cross entropy loss function as baseline model, and compare ribbon charts among ground truth labels, baseline model, and our proposed Meta-SurDiff. The results verify the capability of Meta-SurDiff in reliable OSP recognition, as shown in Fig. \ref{fig:vis} (a) and (b).
Taking predicted ribbon chart on Cholec80 dataset for example, the baseline model easily misclassifies \texttt{P1} (CalotTriangleDissection) into \texttt{P2} (framepingCutting) at middle of the video. The surgical phase name can be found in Appendix \ref{details-data}. In contrast, our model effectively avoid such error.

\noindent \textbf{(2) The learned weight vectors.}
On the other hand, we also visualize the learned weight vectors of 100 training frames uniformly sampled from each phase and report the results in Fig. \ref{fig:vis} (c). 
We find that the learned weights for minority phases in surgical video are typically more prominent than those for majority phases, prompting the model to focus more on frames from minority phases and thereby reducing the risk of overfitting to majority phases.

\section{Conclusion}
\label{sec:conclusion}

We propose Meta-SurDiff for reliable OSP recognition to address two long-overlooked issues 
: i) frame ambiguity in surgical videos; ii) unbalanced frames among different phases. Meta-SurDiff uses a classification diffusion model to calibrate the phase representations for mitigating coarse recognition caused by frame ambiguity. Additionally, we adopt a re-weighting based meta-learning method to train our model for decreasing the risk of overfitting caused by unbalanced phase distribution.
Empirical results show that Meta-SurDiff outperforms other well-designed methods.

\appendix
\section{More Details on Datasets}
\label{details-data}

\textbf{(1) Cholec80 \citep{twinanda2016endonet}.} It comprises 80 laparoscopic surgical videos, with 7 defined phases annotated by experienced surgeons and an average duration of 39 minutes at 25 fps with resolution either 1920 $\times$1080 or 854 $\times$ 480. 
We split the dataset into the 40 videos for training and the rest for testing follow \citep{jin2017sv}. Showcases are illustrated in Fig. \ref{fig:DatasetSample}.

\noindent \textbf{(2) M2Cai16 \citep{twinanda2016miccai}.} It consists of 41 laparoscopic surgical videos with resolution 1920$\times$1080 that are segmented into 8 phases by expert physicians. 
Following \citep{yi2019hard}, we split the dataset into the 27 videos for training and the 14 for testing. Showcases can be found in Fig. \ref{fig:DatasetSample}.

\noindent \textbf{(3) AutoLaparo \citep{wang2022autolaparo}.} It includes 21 laparoscopic hysterectomy videos, with 7 phases annotated by experienced surgeons and an average video duration of 66 minutes recorded at 25 fps with resolution 1920 $\times$1080. Following\citep{liu2023lovit}, we split the dataset into 10 videos for training, and 7 videos for testing. Showcases are illustrated in Fig. \ref{fig:DatasetSample}.

\noindent \textbf{(4) OphNet \citep{hu2024ophnetlargescalevideobenchmark}.} It includes 2,278 surgical videos (284.8 hours), demonstrating 66 different types of ophthalmic surgeries: 13 types of cataract surgery, 14 types of glaucoma surgery, and 39 types of corneal surgery. There are 102 phases and 150 operations for recognition, detection and prediction tasks. Over 77$\%$ of videos have high-definition resolutions of 1280 $\times$ 720 pixels or higher. To facilitate algorithm evaluation, we selected 7,320 phase instances and 9,795 operation instances according to annotated action boundaries, totaling 51.2 hours. The average duration of trimmed videos is 32 seconds, while untrimmed videos average 337 seconds. We used the latest publicly available annotation files from \citep{hu2024ophnetlargescalevideobenchmark}, dividing 445 videos for training, 144 videos for validation, and 154 videos for testing. These videos include 96 phase categories.

\noindent {\textbf{(5) NurViD\citep{hu2023nurvidlargeexpertlevelvideo}.} It is a comprehensive collection of nursing procedure videos that includes 1,538 videos (144 hours) demonstrating 51 different nursing procedures and 177 actions for recognition and detection tasks. Over 74 $\%$ videos have HD resolutions of 1280 $\times$ 720 pixels or higher. To facilitate the evaluation of algorithms, it trimmed the videos based on annotated action boundaries, resulting in 5,608 trimmed video instances that totaled 50 hours and the trimmed videos have an average duration of 32 seconds. Following \citep{hu2023nurvidlargeexpertlevelvideo}, we allocated 3,906 videos for training, 587 videos for validation, and 1,122 videos for testing. Showcases are illustrated in Fig. \ref{fig:DatasetSample}.

All videos are subsampled to 1 fps 
, and frames are resized into 250 $\times$ 250. 
The showcases of each dataset shown in Fig. \ref{fig:DatasetSample} illustrate the visual differences among the five surgical video datasets. 
\begin{figure}[t]
\vspace{-2mm}
\begin{center}
\centerline{\includegraphics[width=0.5\linewidth]{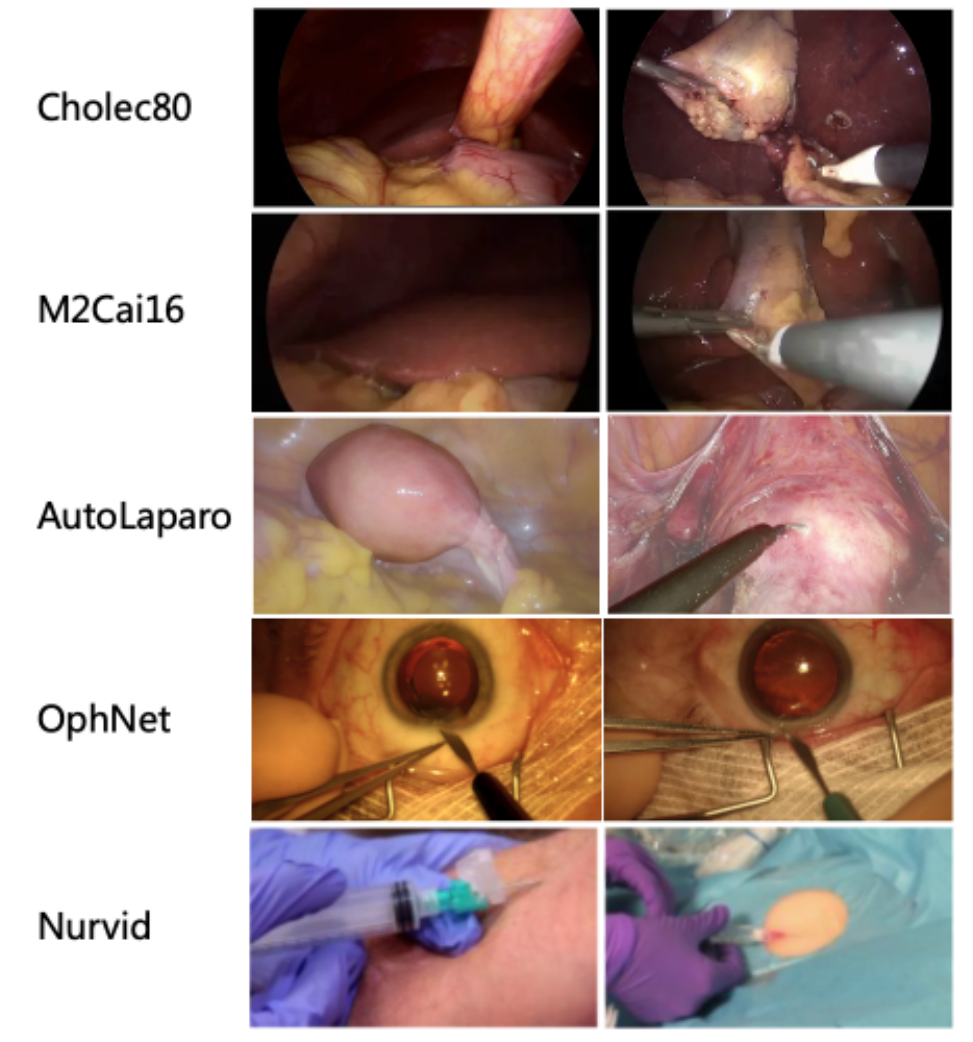}}
\caption{Showcases of surgical video frame for the five datasets. Cholec80, M2Cai16, and AutoLaparo are laparoscopic surgery video datasets, with Cholec80 and M2Cai16 focusing on cholecystectomy, while AutoLaparo is centered on hysterectomy. OphNet is a large-scale dataset for ophthalmic surgery videos, and NurVid is a video dataset for nursing procedure activities.
}
\label{fig:DatasetSample}
\end{center}
\vspace{-8mm}
\end{figure}

\section{More Details on Architectures}
\label{archi}
In our experiments, we set the number of timesteps as $T=100$ and employed a linear noise schedule with $\beta_1=$1e-4 and $\beta_t=0.02$. 
We use a one-layer LSTM, each layer has 512 hidden units. 
As for the learnable projector $g(\cdot)$ in Eq. \ref{eq: condition}, we use a fully-connected layer activated with Softmax function to map the vector from LSTM to the coarse phase representations, also known as conditional inputs $\boldsymbol{z}^{i}$ for $i=1,...,L$ in our work.
As for classification diffusion model, we use a linear embedding for the timestep. Then we concatenate $\boldsymbol{y}^i_{t}$ and $\boldsymbol{z}^i$ altogether and feed them into a three-layer MLP, each with an output dimension of 512. We conduct a Hadamard product between the output vector of the first layer of the MLP and the corresponding timestep embedding, followed by a Softplus nonlinear function.

\section{Assessing Uncertainty with PIW and PTST
}
\label{piw-ttest}
\textcolor{black}{
In OSP recognition, modeling uncertainty holds paramount importance as it directly affects our ability to assess the reliability of decisions for downstream applications, such as real-time monitoring of surgical procedures and optimizing surgeons schedules, which is highly related to the physical health of the people. However, most of the recently developed deterministic methods can not do this job well. In this paper, we introduce Meta-SurDiff and take full advantages of deep generative model to obtain precise frame-level phase distribution. On one hand, uncertainties caused by frame quality and frame distribution can be well minimized at training time by assigning the predicted phase representations with their corresponding ground-truth label embeddings. On the other hand, we could access such uncertainties at test time by generating multiple trajectories given the input frame.
}
To this end, following \citep{fan2021contextual}, we use Prediction Interval Width (PIW) and Paired Two-Sample t-Test (PTST) to help us asessing uncertainty.

\noindent \textbf{(1) Prediction Interval Width (PIW).} It is a statistical measure to assess the uncertainty of prediction results. PIW indicates the width of a certain $\tau \cdot 100$ percent PI, for any value of $\tau \in (0, 1)$:
\begin{align}
\begin{split}
    {\rm{PIW}} (\tau) = \frac{1}{n} \sum_{i=1}^{n} (u_i - l_i), 
    \\ u_i = q_{(1+\tau)/2}^i,  \quad l_i = q^{i}_{(1-\tau)/2}
\end{split}
\end{align}
where $l_i$ and $u_i$ are the lower and the upper bounding quantiles that together define a $\tau \cdot 100$ per cent PI, $q^i_{(1-\tau)/2}$ and $q^i_{(1+\tau)/2}$ are the $(1-\tau)/2$ and $(1+\tau)/2$ quantiles of the predictive distribution of $y^i$. Lower PIW values imply higher sharpness with less uncertainty, which are preferred.

\noindent \textbf{(2) Paired Two-Sample t-Test (PTST).} For single-label classification model, the probabilities are dependent between two classes due to the softmax layer, therefore, we use the PTST, which is an inferential statistical test, to determine whether there is a statistically significant difference between the means of two groups. To obtain the PTST, we calculate the difference between paired observations, often referred to as the t-statistic  $T=\frac{\overline{Y}}{s/\sqrt{N}}$, where $\overline{Y}$ is the mean difference between paired observations, $s$ stands for the standard deviation of the differences, and $N$ is the number of observations. We use this t-statistic and the t-distribution to determine the corresponding p-value.

According to the aforementioned descriptions,
we provide the mean PIW among correct and incorrect predictions, and the mean accuracy among instances rejected and not-rejected by the PTST for all test instances. The details of their calculation are depicted as follows:

\noindent \textbf{(3) The details of calculating PIW and PTST.} For clarity, we set $n$ as the number of instances in the test dataset and $m$ as the number of samples taken for uncertainty assessment for each instance. The data for all instances in the test dataset is represented as $\boldsymbol{x} \in \mathbb{R}^{n \times I}$, where $I$ is feature dimension. Their corresponding ground-truth label with one-hot embeddings are denoted by $\boldsymbol{y} \in \mathbb{R}^{n \times C}$, where $C$ is phase number. 
At test time, each input instance is used to generated $m$ denoised phase representations  $\hat{\boldsymbol{y}} \in \mathbb{R}^{n \times m \times C}$ by reverse diffusion process.
By taking the maximum predicted probability among the $m$ representations for each instance, we obtain the predicted recognition results $\overline{\boldsymbol{y}} \in \mathbb{R}^{n \times C}$ for all instances. 
Furthermore, extracting the two highest probabilities from the $m$ predictions yields two distributions, each containing $m$ values for each instance. These two distributions serve as the two input samples for PTST statistical testing, denoted collectively as $t \in \mathbb{R}^{n \times m \times 2}$. Comparing the predicted results $\overline{\boldsymbol{y}}$ with the ground-truth label embeddings $\boldsymbol{y}$ allows us to determine the number of correctly classified and misclassified samples, thereby the recognition accuracy is obtained. Subsequently, we compute the PIW values between the 2.5th and 97.5th percentiles of the predicted probabilities separately for correct and incorrect samples, averaging these values to derive the PIW for correct and incorrect samples. Besides, we conduct hypothesis tests on the obtained dual-sample distribution $t$, yielding a p-value for each instance, which is compared against the given significance level $\rho=0.05$ to classify instances as 'reject' or 'no-reject' groups. It's noteworthy that these hypothesis tests are based on the largest and second largest predicted probabilities for each instance. Therefore, the hypothesis is that "the top two maximum predicted values are the same". Larger differences between these probabilities indicate higher accuracy in the maximum predicted probability, identifying 'reject' samples as those with better predictive performance. Finally, we tally the number of correctly predicted samples in the 'reject' and 'no-reject' groups, calculating the corresponding accuracy for each group.

\bibliographystyle{plainnat}
\bibliography{main}

\end{document}